\documentclass[twoside,11pt]{article}
\usepackage[T1]{fontenc}  
\usepackage{jair, theapa, rawfonts}
\jairheading{}{2024}{1-32}{6/27}{---}
\firstpageno{1}
\ShortHeadings{Defining and Evaluating AI Group Fairness: a Fuzzy Logic Perspective}{Krasanakis \& Papadopoulos}

\usepackage{graphicx} 
\usepackage{amsmath}
\usepackage{amssymb}
\usepackage{amsfonts}
\usepackage{mathbbol}
\usepackage{relsize}
\usepackage{amsthm}
\usepackage{url}
\DeclareMathOperator{\prob}{\mathcal{P}}
\newtheorem{theorem}{Theorem}

\newtheorem{proposition}{Proposition}

\begin{document}

\title{Evaluating AI Group Fairness: a Fuzzy Logic Perspective}
\ShortHeadings{Evaluating AI Group Fairness: a Fuzzy Logic Perspective}
{Krasanakis \& Papadopoulos}
\author{\name Emmanouil Krasanakis \email maniospas@iti.gr
\\\addr Centre for Research \& Technology Hellas, Information Technologies Institute, 57001, Greece
\AND \name Symeon Papadopoulos \email papadop@iti.gr \\\addr Centre for Research \& Technology Hellas, Information Technologies Institute, 57001, Greece}

\maketitle

\begin{abstract}
Artificial intelligence systems often address fairness concerns by evaluating and mitigating measures of group discrimination, for example that indicate biases against certain genders or races. However, what constitutes group fairness depends on who is asked and the social context, whereas definitions are often relaxed to accept small deviations from the statistical constraints they set out to impose. Here we decouple definitions of group fairness both from the context and from relaxation-related uncertainty by expressing them in the axiomatic system of Basic fuzzy Logic (BL) with loosely understood predicates, like encountering group members. We then evaluate the definitions in subclasses of BL, such as Product or Lukasiewicz logics. Evaluation produces continuous instead of binary truth values by choosing the logic subclass and truth values for predicates that reflect uncertain context-specific beliefs, such as stakeholder opinions gathered through questionnaires. Internally, it follows logic-specific rules to compute the truth values of definitions. We show that commonly held propositions standardize the resulting mathematical formulas and we transcribe logic and truth value choices to layperson terms, so that anyone can answer them. We also use our framework to study several literature definitions of algorithmic fairness, for which we rationalize previous expedient practices that are non-probabilistic and show how to re-interpret their formulas and parameters in new contexts.
\end{abstract}

\section{Introduction}
Artificial Intelligence (AI) systems see pervasive use in many applications, where they automate predictive tasks affecting people's lives, like classification and recommendation. Given the tendency of machine learning to replicate and exacerbate real-world discrimination, making AI fair is therefore a subject of intensive research \cite{ntoutsi2020bias,barocas2023fairness,mitchell2021algorithmic,mehrabi2021survey}. 
Defining fairness depends on the social context, such as the geographic area, local history, and culture in which systems are deployed. It also depends on the stakeholders expressing fairness concerns \cite{john2022reality,li2023trustworthy}, such as domain experts, policymakers, business owners, or underrepresented groups affected by the systems. 

Definitions of fairness tend to echo philosophical statements \cite{binns2018fairness} and include but are not limited to looking at measures of discrimination\footnote{Example measures of discrimination are presented in Section~\ref{measures}. We do not call these measures of bias, due to separating the two terms in Section~\ref{propositions}.} that capture context-specific concerns. For example, \textit{disparate impact} concerns that equalize positive rates between different values of protected attributes, like gender (men vs women) and race (Caucasian vs African vs Asian, etc.), originate from the context of US regulation \cite{barocas2016big}. Measures of discrimination are optimized by integrating them in algorithms, for instance as constraints or training objectives, during data pre-processing, or during post-processing of predictions. In this work, we focus on definitions of \textit{group fairness}, which capture predictive disparities between groups of people. These often matter when they reflect differences between demographics, such as genders, races, or their intersections.

Group fairness commonly combines statistical reasoning about properties of inter-group parity with relaxations that are necessary in practice (Subsection~\ref{justification}). For example, numerical properties may be considered similar between groups if their relative differences are small. Relaxations introduce free choices, like the maximum accepted relative difference, that are rarely accompanied by logical interpretation and are thus hard to determine in practice. Take, for example, the concern that a binary classifier may exhibit \textit{disparate mistreatment} between men and women by producing different false positive rates between these two genders \cite{zafar2017fairness}. Since equalizing the rates may require predictive performance concessions, definitions of fairness often accept a relative rate difference of at most 20\%.\footnote{For example, the 20\% threshold is recommended by the online application of the Aequitas \cite{saleiro2018aequitas} framework  (\url{http://aequitas.dssg.io/upload.html}).} This threshold is recommended in guidelines addressing the \textit{different problem and -often- context} of \textit{disparate impact} between men and women \cite{biddle2017adverse}, where the relative difference is measured between the positive rates of these genders. However, the threshold's usage for disparate mistreatment or for the protection of attributes other than gender remains unjustified.

In general, intuitive terminology is needed to justify fairness definitions with a mixture of philosophical concerns, regulation, and stakeholder beliefs \cite{weinberg2022rethinking}. To this end, we argue that fairness can be first defined axiomatically and be adapted across different contexts afterwards. Axiomatic definitions would be expressed through abstract predicates and logical connectives, like conjunction (two statements holding true simultaneously) and implication. Such definitions already exist for predicates with binary truth values \cite{ignatiev2018sat,ignatiev2020towards,belle2023toward,lee2023fair}. However, we recognize that some predicates may correspond to uncertain beliefs or continuous quantities, such social values \cite{cheng2021socially} and measures of discrimination that capture inter-group disparity (Section~\ref{demonstration}). Additionally, they may not be perfectly understood a-priori or could admit a different interpretation in each context, depending on measured physical quantities and stakeholder opinions \cite{mielke2016stakeholder,wachter2021fairness}.

To evaluate group fairness that depends on uncertain predicates while leveraging the inherent explainability of logical statements, we resort to fuzzy logic. This generalizes probabilities to truth values in the range $[0,1]$, but is not necessarily derived from a statistical measurement of reality. Thus, it can model any kind of continuous evaluation or belief system that changes across contexts. In Section~\ref{our approach} we introduce a framework in which researchers, AI creators, policy makers, or logic practitioners define group fairness and corresponding bias through abstract predicates in Basic fuzzy Logic (BL); this is a class of fuzzy logic variations that share the same base axioms but model different truth value evaluation mechanisms. To simplify adoption of our framework by removing the need to reason in BL (which is not widely known or easy to learn) we create a standardization of what group fairness evaluation should look like under commonly accepted propositions.

In our analysis, fairness or bias definitions expressed in BL can be evaluated to their continuous truth values based on three choices that differ between contexts: i) the logic subclass to work in, which determines the evaluation mechanism, ii) how the truth value of discrimination is computed, such as using established measures of discrimination from the literature, and iii) the truth value of the predicate ``group membership''. We also explain how these choices can be transcribed to intuitive terms so that they can be retrieved from stakeholder opinions. We finally map them to existing algorithmic practices in certain situations to bring intuitive understanding and the option of modeling different belief values with the same reasoning. Our contribution is threefold:
\begin{itemize}
    \item[a)] We introduce a framework that integrates stakeholder beliefs in formal definitions of fairness written in BL. This bridges the gap between axiomatic rigor and the context-dependent nuance of fairness; we allow expressing high-level statements independently from capturing or learning to replicate the truth values of beliefs. 
    \item[b)] We standardize definitions of group fairness and bias for common propositions stated in BL. In particular, we derive closed-form equations where the chosen logic's evaluation mechanism for connectives and the truth values of predicates may be plugged in to numerically quantify fairness.
    \item[c)] We use our framework to study several ad hoc fairness evaluation strategies found in the literature. We identify generating BL definitions, the logic type, and which predicates mirror fairness parameters, such as thresholds. Thus, definitions can be adjusted to new contexts or beliefs, and we demonstrate such alternatives.
\end{itemize}

This paper is structured as follows: Section~\ref{background} provides background on group fairness, and fuzzy logic that models a broad range of axiomatic systems supported by our analysis. Section~\ref{our approach} overviews our proposed fairness definition framework and investigates commonly held propositions about the terms discrimination, imbalance, bias, and fairness between groups of people. Starting from the propositions, it also analytically standardizes definitions of group fairness and corresponding bias within BL, and presents intuitive interpretation of free choices that can be quantified through stakeholder beliefs. Section~\ref{demonstration} applies the framework on four existing definitions of fairness, where we uncover interpretations of expedient practices in non-probabilistic logics and suggest valid alternatives for different contexts. Section~\ref{challenges} discusses practical adoption through an interdisciplinary workflow, and presents related challenges. Finally, Section~\ref{conclusions} summarizes and concludes our work.

\section{Background}\label{background}
Here we present the necessary background to understand our analysis. In Subsection~\ref{measures} we summarize related literature on group discrimination. In Subsection~\ref{possibilistic} we provide a brief introduction on fuzzy logic and its subclasses we work with. A certain degree of familiarity with them helps AI system creators adopt our framework, but is not needed by the stakeholders, who only state abstract definitions of fairness (these definitions are transcribed to BL expression by system creators---Section~\ref{challenges}) or express belief values.

\subsection{Related work}\label{measures}
Defining and imposing fairness on AI systems is the subject of ongoing debate, which often considers many types of mathematical and procedural evaluation, like the EU's Assessment List for Trustworthy AI \cite{ala2020assessment}, and NIST's AI Risk Management Framework \cite{ai2023artificial}. Among other actions, fairness also accounts for discrimination in data or predictions with mathematical definitions \cite{ntoutsi2020bias,barocas2023fairness,mitchell2021algorithmic,mehrabi2021survey}. These definitions can be broadly categorized into i) group fairness that focuses on equal treatment between population groups or subgroups (see below), and ii) individual fairness that focuses on fair treatment of individuals, for example by yielding similar predictions for those with similar features  \cite{dwork2012fairness}. Making fair AI typically involves measures of discrimination that quantify the numerical deviation from exact definitions of fairness. These are either minimized or subjected to constraints \cite{xinying2023guide}.

We now present common measures of discrimination tied to group fairness concerns of classifiers \cite{castelnovo2021zoo}. These aggregate comparisons between groups of data samples, and analogous ones port the same underlying principles to other predictive tasks \cite{li2023trustworthy,xinying2023guide}. The measures of $prule$ \cite{biddle2017adverse} and Calders-Verwer disparity $cv$ \cite{calders2010three} quantify prediction rate equality between two groups of samples $\mathcal{S}$ and $\mathcal{S}'$, where the latter may complement the former. Mathematically:
$$prule=\min\Big\{\frac{\prob(c(x)=1|x\in \mathcal{S})}{\prob(c(x)=1|x\in \mathcal{S}')},\frac{\prob(c(x)=1|x\in \mathcal{S}')}{\prob(c(x)=1|x\in \mathcal{S})}\Big\}$$
$$cv=|\prob(c(x)=1|x\in \mathcal{S})-\prob(c(x)=1|x\in \mathcal{S}')|$$
where $\prob(\cdot|\cdot)$ denotes conditional probability and $c(x)\in\{0,1\}$ the binary outcome of classifying data sample $x$. The disparate impact concern is eliminated when differential fairness for each group intersection $df=1-prule$ \cite{foulds2020intersectional}, or equivalently $cv$, becomes $0$. 
Another concern is disparate mistreatment \cite{zafar2017fairness}, which captures misclassification differences between groups as:
$$|\Delta m|=|m(\mathcal{S})-m(\mathcal{S}')|$$
where $m(\cdot)$ is a misclassification measure over a group of samples, such as their false positive rate (fpr) or their false negative rate (fnr). The same formula can also express $cv$ or error measures for other predictive tasks, and therefore constitutes a building block of generalized fairness evaluation \cite{roy2023multi}. An example of how difference-based discrimination may look for probabilistic measures of predictive performance is the absolute value of equalized odd differences \cite{hardt2016equality,donini2018empirical}:
$$|\Delta eo|=|\prob(c(x)=1|x\in \mathcal{S}, Y(x)=y)-\prob(c(x)=1|x\in \mathcal{S}',Y(x)=y)|$$
where $y\in\{0,1\}$ and $Y(x)$ is the true test/validation data label corresponding to sample $x$.
In more complex scenarios, fairness concerns may span several sensitive groups and measures of discrimination. For example, there may be multiple protected demographics (e.g., genders, races, and their intersections), whereas disparate mistreatment of classifiers may be quantified as $|\Delta fpr|+|\Delta fnr|$ \cite{krasanakis2018adaptive}. Multiple concerns could be contradictory \cite{kleinberg2016inherent}, but corresponding discrimination could still be aggregated to one quantity \cite{hooker2012combining,roy2023multi}, for instance through a maximum, minimum, average, or weighted average. Multiple concerns can also model individual fairness \cite{dwork2012fairness,kim2019preference}, for example by setting each individual as a separate group. Reductions generally stem from numerical intuition, but in Subsection~\ref{multi} we justify some of them with fuzzy axiomatization.

Creating logically consistent -and interpretable- fairness definitions is an emerging literature concern. First, statistical tests have been proposed as alternatives to other measures of discrimination \cite{watkins2022four}, like the $prule$ or $cv$, by rejecting the null hypothesis that group \textit{distributions} (instead of aggregate quantities) are indistinguishable. Such tests also arbitrarily decide when results are statistically significant; under our framing, p-values can serve as measures of discrimination to let stakeholders decide which confidence levels matter. Similarly, causal models have been proposed as a means to learn parameterized discrimination mechanisms and make predictions in a would-be fair reality, an approach called counterfactual fairness \cite{kusner2017counterfactual,carey2022causal}.\footnote{Counterfactual fairness is often mentioned as a standalone approach or in relation to individual fairness, but Rosenblatt and Witter \cite{rosenblatt2023counterfactual} observe that it is a type of group fairness.} Causal modeling heavily relies on true world models, inaccuracies of which also set counterfactual fairness as an approximation. Recent variations further suggest sequences of human actions instead of only generating explainable models \cite{karimi2021algorithmic,verma2020counterfactual,depersonalized}; obtaining measures of discrimination about suggested actions remains an open question that our framework could help address. 

Fairness definitions have also been epistemized, that is, transcribed from abstract statements to unambiguous mathematical expressions, within propositional logic \cite{ignatiev2018sat,ignatiev2020towards,belle2023toward,lee2023fair}. These fail to support practical relaxation into approximate measures of discrimination, and instead tend to coarsen continuous-valued intermediate assessments to binary truth values. Alternatively, they impose \textit{exact} fairness while maximizing accuracy, which can create extreme practical trade-offs. In terms of axiomatizing fairness uncertainty, initial links between Rawl's difference principle \cite{ashrafian2023engineering} and the logical system of G{\"o}del fuzzy logic have been identified \cite{suzuki2018axiomatic}. Finally, belief merging \cite{konieczny2002merging}, which refers to combining different truth values of beliefs, has been used on the same class of t-norm logic that we explore \cite{oliveira2018propositional} to derive principles similar to Subsection~\ref{multi}, but has not been applied on merging beliefs about fairness definitions that would require logical consistency with fairness axiomatization.

\subsection{Fuzzy Logic and BL}\label{possibilistic}
Fuzzy logic facilitates reasoning over uncertain facts by introducing predicates $X$ with truth values in the unit interval $\prob(X)\in[0,1]$ \cite{zadeh1978fuzzy}. 
Truth values are not necessarily probabilities, but we adopt the same symbol because they can model probabilities in some logic. When needed, we use subscripts to refer to specific logic variations; for example, $\prob_L(\cdot)$ computes truth values under logic $L$. 

Recent efforts axiomatize how truth values can be evaluated for logical expressions given the respective truth values of involved predicates and some definitions of logical connectives. Connectives used in this work are conjunction $\&$, negation $\neg$, implication $\to$, and idempotent conjunction $\wedge$. 
Axiomatization efforts are centered around Monoidal T-norm based Logics (MTLs) \cite{esteva2001monoidal,dubois2004possibilistic,spada2008short,sep-logic-fuzzy}, where numerical operators $\star$ called t-norms evaluate the truth value of conjunction, that is, of two expressions simultaneously being true:
$$\prob(X\,\&\,Y)\triangleq\prob(X)\star\prob(Y)$$
where $\triangleq$ denotes an equality that holds by definition and has lower priority than all other symbols. All t-norms should satisfy the properties described in Appendix~\ref{tnorms}, as do the ones presented in Table~\ref{tab:t-norms}. The satisfied properties lead to the existence of a numerical operator $\Rightarrow$ called the residuum, that models the truth value of implication:
$$\prob(X\to Y)\triangleq\prob(X)\Rightarrow \prob(Y)$$
Both t-norms and residua are closed-form numeric functions chosen together with the corresponding logic. That is, they present specific evaluation formulas. As a computational example, $0.1\star_\Pi 0.2=0.02$ and $0.2\Rightarrow_\Pi 0.1=0.5$ in Product logic ($\Pi$). To avoid confusion, we use the residuum symbol for studied logics, and textual statements for our reasoning.

Finally, each logic defines what is called a strong negation of expressions $X$ based on mutual exclusiveness: $$\overset{str}\neg X\triangleq X\to False\quad\text{for which}\quad X\,\&\,\overset{str}\neg X=False$$ where $False$ is the logic's fallacy predicate with zero truth value $\prob(False)=0$. Fuzzy set theory can be modeled by also adding a weak negation predicate $\overset{wk}{\neg}$, which corresponds to conceptually opposite but not necessarily contradictory terms. For example, this operator can model set complements, where elements may belong both to a set and to its complement with complementary truth values. Weak negation is defined through its truth value:
$$\prob(\overset{wk}{\neg} X)=1-\prob(X)$$

BL \cite{hajek1998basic,cignoli2000basic} is a class of MTLs whose members share a fixed set of axioms but each corresponds to different continuous t-norms.\footnote{Generally, t-norms need only be left-continuous. But any fully continuous one can model BL.} It is easier to work with due to additional properties. Subclasses of BL refer to logics that adopt specific norms to create a unique evaluation mechanism. This is equivalent to admitting certain additional axioms described in Appendix~\ref{axiomatization}. \textit{Any property holding for BL holds for any of its subclasses}. In this logic, idempotent conjunction is defined per:
$$X\wedge Y\triangleq X\,\&\,(X\to Y)\quad\text{with truth value}\quad\mathcal{P}(X\wedge Y)=\min\{\mathcal{P}(X),\mathcal{P}(Y)\}$$ This is stronger than non-idempotent conjunction in that the expression $(X\,\&\,Y)\to(X\wedge Y)$ is a tautology (always has truth value 1) in any BL. 

The Mostert-Shields theorem \cite{klement2013triangular} states that all continuous t-norms are locally isomorphic to the three types in Table~\ref{tab:t-norms}. If all truth values $x \not\in\{0,1\}$ are nilpotent, i.e., satisfy $x\star x\star\dots x=0$ for some number of conjunctions, t-norms are isomorphic to Lukasiewicz logic. This has the same strong and weak negation, and it fully models propositional logic \cite{chang1959new}.
Otherwise, if t-norms are indempodent, i.e., $x\star x=x$, they are isomorphic to Godel logic, which avoids degradation of truth values as more propositions are conjuncted. Finally, other t-norms that do not exhibit the above two properties for every $x\in[0,1]$ are isomorphic to Product logic, which can model probability theory \cite{hajek2013fuzzy,flaminio2018towards,yurchenko2021foundations}.

\begin{table}[htpb]
\small
    \centering
    \begin{tabular}{l l l}
        \textbf{Logic} & \textbf{T-norm} & \textbf{Implication}\\
        \hline
         G{\"o}del &  $x\star_G y\triangleq \min\{x,y\}$ & $x\Rightarrow_G y\triangleq \{1\text{ if }x\leq y, y\text{ otherwise}\}$\\
         Product &  $x\star_\Pi y\triangleq  x\cdot y$ & $x\Rightarrow_\Pi y\triangleq \{1\text{ if }x\leq y, \tfrac{y}{x}\text{ otherwise}\}$\\
         Lukasiewicz &  $x\star_L y\triangleq \min\{x+y-1,0\}$ & $x\Rightarrow_L y\triangleq \min\{1,1-x+y\}$\\
    \end{tabular}
    \caption{Base BL subclasses.}
    \label{tab:t-norms}
\end{table}

\section{Evaluating Group Fairness in Fuzzy Logic}\label{our approach}
\begin{figure}[htpb]
\centering
\includegraphics[width=0.8\textwidth]{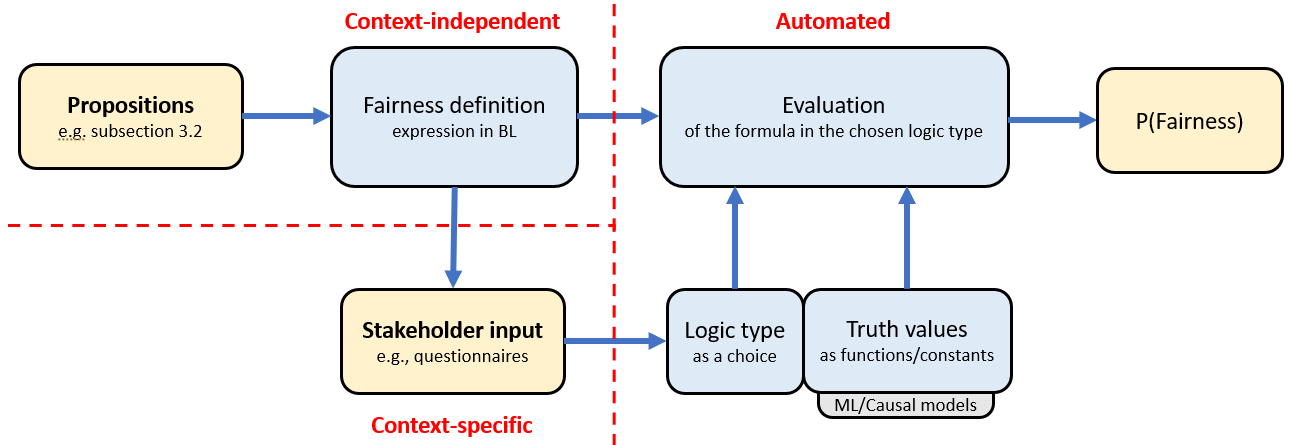}
\caption{Our fuzzy framework for evaluating fairness. Shaded cells denote inputs/outputs.}\label{overview}
\end{figure}

In this section we introduce a fairness evaluation framework that is based on fuzzy logic and summarized in Figure~\ref{overview}. First, definitions of fairness are written in BL using abstract predicates, such as bias and discrimination, that may obtain non-binary truth values. The definitions could be derived from commonly held base concepts like we do in Subsection~\ref{propositions}, from policy makers, or from regulation that is emerging worldwide. Exact interpretation of predicates could still be ambiguous at this stage; BL admits the axioms of Appendix~\ref{axiomatization} to support inference, if needed, but clarifying interpretations to obtain specific truth values will depend on the context too. In other words, definitions transcribe context-independent concerns into formal language. Practical adoption of our framework can reimagine any of the propositions and definitions we later provide, which mainly serve as starting points that let AI system creators bypass common analysis.

Evaluating the truth value of fairness also requires stakeholder feedback to determine the logic type to work with (the BL subclass) and the truth values of abstract predicates. Stakeholders do not need to look at expressions involving logical connectives, and can be consulted through social science processes, such as questionnaires about the degree of group membership and discrimination in certain scenarios. Deep learning \cite{lecun2015deep} or causal models \cite{petersen2014causal,gopnik2012reconstructing,rohrer2018thinking} may use example answers to learn functions that compute truth values from available data. Finally, truth values are plugged into fairness definitions to evaluate them, namely produce the context-specific truth value $\prob(Fair)$.
This can be combined with additional predicates to evaluate increasingly complex dependent expressions, such as the multidimensional and individual fairness concerns showcased in Section~\ref{demonstration}. The framework constitutes a generalization of counterfactual fairness definitions to include non-probabilistic logics in the modelled implication mechanism, and can analyze non-group fairness too.

Subsection~\ref{justification} explains why a purely probabilistic approach may fail to support practical implementations of group fairness with intuitive explanations. 
Subsection~\ref{propositions} introduces a set of logical predicates that are involved in defining fairness throughout this work. It also presents propositions that typically apply on these predicates, which we use in Subsection~\ref{standardize} to create standardized forms of evaluating group bias and fairness formulas in BL. This makes it easier to create new definitions in practice without needing to reason in that logic other than making generic statements that are directly converted to numerical expressions. Finally, Subsection~\ref{logictype} explains how choices needed for context-specific fairness evaluation can be guided by stakeholder input. Gathering rigorous input is left to future work.

\subsection{Why not stick to probabilities?}\label{justification}
Most measures of group discrimination start by recognizing probabilistic definitions of fairness, usually equating base AI assessment measures between groups or subgroups. However, what regularly follows next is a violation of logical consistency that a) evaluates discrimination through arbitrary continuous numerical comparisons of the base assessments, such as differences, and b) imposes ad hoc criteria of what constitutes ``small enough'' discrimination to achieve fairness. Even practices that consciously strive to maintain probabilistic reasoning, such as differential fairness or statistical tests (Subsection~\ref{measures}), find themselves accepting thresholds that may be hard to justify, especially when transferred between contexts without consultation. Thus, they lean on theorization outside of probabilistic justification.

Relaxing exact definitions of bias or fairness could be necessary in practice, especially when creating tractable implementations or optimization goals, such as differentiable losses or counterfactual model parameters to be learned. The end result is that the definitions being evaluated do not exactly satisfy the theoretical properties they set out to impose. This could be avoided given some intuitive interpretation of the relaxations. That is, perfect tightness represents the ideal fair scenario, but what non-perfect tightness entails should also be understood. If such an understanding is achieved, we argue that relaxations do not necessarily violate logical consistency, but may instead model slightly different fairness definitions that arise from expression and predicates stated in axiomatic systems like BL. Recall that the latter is broader than pure probabilistic thinking, and therefore accommodates more predicates, for example that are evaluated based on stakeholder beliefs.

\subsection{Terminology and propositions}\label{propositions}
\begin{figure}[htpb]
\centering
\includegraphics[width=0.7\textwidth]{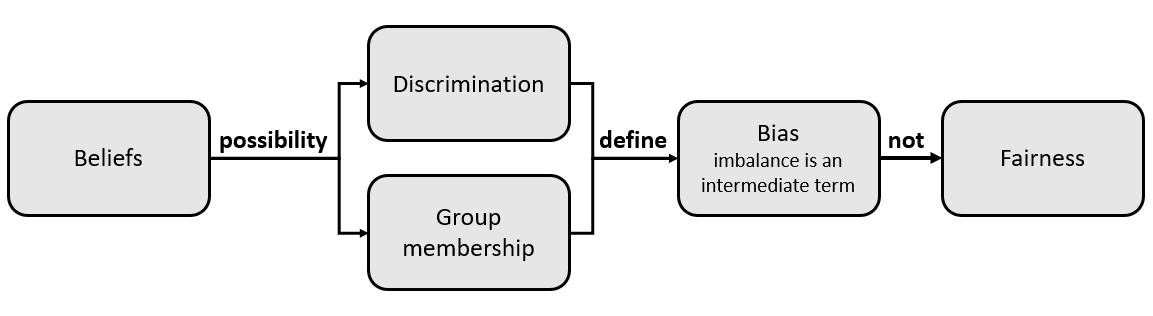}
\caption{Relation between terms employed in our fairness propositions.}\label{flow}
\end{figure}

Throughout this work, we separate the terms \textit{discrimination}, \textit{imbalance}, \textit{bias}, and \textit{fairness}. Contrary to when the first three are used interchangeably, we establish them as logical predicates that do not always coincide. Broadly, we refer to discrimination as properties that differ between groups of data samples (e.g., protected groups against all the other samples) and its truth values correspond to what are commonly known as measures of bias, which in this work are renamed to \textit{measures of discrimination}. Our analysis starts from already established measures of this kind, and parses the discrimination predicate they evaluate with fuzzy logic. Imbalance captures that a certain degree of discrimination implied by group membership, and bias captures imbalance for groups. Finally, fairness indicates the absence of bias. Term relations are presented in Figure~\ref{flow} and detailed below.

We introduce BL propositions that mirror commonly-held assumptions about the above terms. These cover a wide range of fairness definitions, many of which are shown in Subsection~\ref{standardize} to be convertible to well-structured evaluation formulas that we call their \textit{standard form}. Different propositions could be stated in BL or other types of fuzzy logic and plugged in our generic framework. That said, this first analysis we conduct maps to several established practices and can be applied out-of-the-box in many systems and contexts, as demonstrated in Section~\ref{demonstration}.

We base our propositions/definitions on two predicates; \textit{discrimination} $E$, which we set as a generic property that holds over some system or set of predictions, and \textit{group membership} $S$ for a specific group of people, for example to be protected. Based on practical needs, this could could be any examined group, subgroup, or even individual (group of one person). Exact interpretation of both predicates is vague and their truth values may change across contexts. For example, group membership may correspond to the possibility of making predictions for a group's member. Later in this subsection we also present an optional proposition that could derive discrimination from simpler concepts. 

We now present propositions that are high-level statements whose truth values heavily depend on the context-specific interpretation of predicates. Their implicit use in the literature can be verified by checking whether they can replicate existing practices, as happens in Subsection~\ref{replicating}. We start with a term that we call \textit{imbalance} that describes how someone may infer discrimination based on group membership. Thus, it should logically imply the discrimination predicate, and also be defined as an implication from group membership to some property. When no group members are encountered, we consider imbalance to be fully certain (have a truth value of $1$). In what we later call worst-case bias, imbalance captures the principle of group membership leading to discrimination against members of the group.

\begin{proposition}\label{imbalance}
Group imbalance $Imb$ is an implication from group membership to some property such that $\prob(Imb)=1$ for no group members and $Imb\to E$ otherwise. 
\end{proposition}

We next introduce the concept of \textit{group bias} corresponding to some definition of group fairness as a conjunction between imbalance and group membership. This essentially captures the event of encountering group members while imbalance occurs. Notice that the intermediate term of imbalance lets us simplify both this proposition and the theoretical analysis of Subsection~\ref{standardize}, and is a weaker common denominator of discrimination and bias that is continuous-valued almost everywhere (except from $s=0$) for BL subclasses.

\begin{proposition}\label{group bias}
Group bias is imbalance that occurs for group members:
\begin{equation}Bias\triangleq  S\,\&\,Imb\end{equation}
\end{proposition}

Finally, we introduce the logical predicate $Fair$ that represents group fairness as the strong logical negation of bias to make the two mutually exclusive. Alternative definitions that would induce non-binary fairness values for G{\"o}del or Product logic, such as the \textit{Unbias} of Subsection~\ref{multi} that captures the weak negation of bias, would not satisfy mutual exclusivity. Therefore, they would create artificially easy routes for improving fairness that are not logically inconsistent. Instead, we argue that additional propositions are needed to be perfectly fair in certain logics, as we demonstrate in Subsection~\ref{replicating}. Although bias is an intermediate term, its definition tends to be a continuous function of the truth values of base predicates in all logics. Thus, fairness can be achieved by (perfectly) minimizing bias.

\begin{proposition}\label{fairness}
Group fairness occurs when there is no group bias:
\begin{equation}Fair\triangleq  \neg Bias\end{equation}
\end{proposition}

The above propositions rely on a base discrimination predicate. This can either be defined through stakeholder input, or derived from further axiomatization. The latter option is often chosen in the literature by checking whether groups differ over some base property $M(S)$. This is captured in a final proposition, where groups $S$ are said to be indistinguishable from different groups $S'$ (the latter could be their complement in the population, but not necessarily) when the property holding for one implies the property holding for the other and conversely. We use the weak negation of indistinguishability because we do not consider the two concepts to be perfectly exclusive; discrimination in this case is a lax term and the rest of the propositions are needed to convert it to an indication of bias or fairness. We stress that closed-form definitions of discrimination are optional, and this concept would ideally be either negotiated or learned from a combination of stakeholder examples and base predicates, as demonstrated in Appendix~\ref{axiomatization}.

\begin{proposition}[optional]\label{discrimination}
Discrimination is the weak negation of indistinguishability:
\begin{equation}\label{discform}E\triangleq\overset{wk}\neg \big(M(S) {\leftrightarrow} M(S')\big)\end{equation}
where two-way implication is defined as $X {\leftrightarrow} Y \triangleq  (X\to Y)\,\&\,(Y\to X)$.
\end{proposition}

\subsection{Evaluating group fairness}\label{standardize}
We now switch to analyzing the truth values of the previous subsection's propositions. We reiterate that, in presented results, t-norms $\star$ and residua $\Rightarrow$ are numeric operations that correspond to the chosen BL subclass (results hold for every subclass). For brevity, we annotate the truth values of the two base predicates as $e=\mathcal{P}(E)$ and $s=\mathcal{P}(\mathcal{S})$, and of imbalance and bias as $imb=\prob(Imb)$ and $bias=\prob(Bias)$ respectively. Numerically, imbalance captures further transformations that measures of discrimination would be subjected to, therefore underestimating discrimination the latter inserted it in a bias evaluation pipelines; when $imb\Rightarrow e$, the properties of the residuum yield $imb\leq e$. 


In Theorem~\ref{exists imbalance} (proof in Appendix~\ref{proofs}) we extract a well-structured BL expression that can be used in lieu of imbalance to replace it in a wide breadth of evaluation scenarios. We call the new expression the \textit{standard form} of imbalance definitions, and use it to standardize subsequent analysis. The standard form is not necessarily an imbalance definition too, but it yields the same final numerical evaluation if replaced in the definition of bias.

\begin{theorem}\label{exists imbalance}
For every imbalance definition $Imb$ there exists expression $F(S,E)$ such that:
\begin{equation}\label{canonical}
    Imb_{F}\triangleq S\to (S\,\&\,E\,\&\,F(S,E))
\end{equation}
satisfies $s\star\prob(Imb_{F})=s\star imb$ as long as: 
\begin{equation}\label{feasibility}
    \prob(F(S,E))\leq \prob(F_{gen}(S,E))\text{ where }\prob(F_{gen}(S,E))\triangleq\Big(e\Rightarrow \inf_{s\star d\geq s\star e} d\Big)
\end{equation}
\end{theorem}

Given the above, Theorem~\ref{Fgroup} (proof in Appendix~\ref{proofs}) extrapolates a generic formula for the truth value of group bias. There is much leeway for customization through the free expression $F(S,E)$ first introduced in the standard form of imbalance. Still, we get a sense for general properties of bias and fairness. For example, the formula for bias does not contain any in-built residuum (which would correspond to implication) and therefore logics isomorphic to Product logic can only rely on the free expression to introduce thresholding criteria via implication; if these criteria are absent, for example when the free expression always has truth value $1$, no valid definition of bias can include thresholds under Product logic. This is a generalization of existing critique against applying thresholds on probabilities.

\begin{theorem}\label{Fgroup}
For a standard form of imbalance $Imb_F$, the corresponding group bias definition has truth value:
\begin{equation}bias_F\triangleq
s\star e\star\prob(F(S,E))\end{equation}
\end{theorem}

In Theorem~\ref{worst bias} (proof in Appendix~\ref{proofs}) we introduce the concept of \textit{worst-case bias} whose truth value is the greatest possible one, even when biases are not be expressed in standard form. The theorem's inequality is trivial to show, even without the rest of our analysis, given that $imb\leq e$ by proposition. However, the inequality is shown to be \textit{strict} in that there exists a standard form of bias that corresponds to a specific predicate selection $F_{gen}(S,E)$, called \textit{worst-case generator}, and achieves the worst case under any BL subclass and any truth values $s,e$. In practice, worst-case bias may be adopted by generic theoretical analysis; when its strong negation yields perfect fairness, the same happens for other bias definition. 

\begin{theorem}\label{worst bias}
Any bias definition in BL has truth value at most:
\begin{equation}bias\leq bias_{F_{gen}}=s\star e\end{equation}
\end{theorem}

As a final remark on the worst-case generator, we use it to understand the limits of which definitions can be expressed in standard form. To this end, notice that Theorem~\ref{Fgroup} can replicate any bias definition arising from any imbalance definition in a standard form when the restriction of Equation~\ref{feasibility} is trivial to meet, that is, when the generator has truth value of $1$ for $e\leq \inf_{s\star d\geq s\star e}d$. We call the set of accepted values $e$ satisfying this condition given a constant $s$ as the \textit{generation space} of the standard form. This depends solely on the choice of the t-norm. Subsection~\ref{replicating} presents the generation spaces of the three base logics.

\subsection{Truth values from stakeholder beliefs}\label{logictype}
Here we explain how stakeholder feedback can determine the context-specific evaluation of bias and fairness, that is, the quantification to truth values of definitions already written in BL. Evaluation requires knowing which BL subclass (characterized by its t-norm) supplies the evaluation mechanism, as well as how to evaluate group membership $S$, discrimination $E$, and the predicate $F(S,E)$ on system outputs. When there is no additional proposition or relevant input, the last predicate can be chosen as its worst-case version to create the strongest definition possible. When Proposition~\ref{discrimination} is adopted, choosing a measure of discrimination is also simplified to choosing a measure of algorithmic performance.

Extracting context-specific truth values for the aforementioned predicates can be achieved either through social negotiation processes to arrive at basic statements and their truth values in a given context, or with deep learning or causal models trained on example situations that have been assessed by stakeholders.\footnote{Even if machine learning models directly quantified bias or fairness, they would be subject to our analysis if they satisfied the same propositions. Thus, given that a certain degree of explainability is key for responsible and fair AI, there is value in retaining an axiomatic system on top of learning processes.} For example, a context-dependent interpretation of $\mathcal{S}$ could be the social separation between a group and the rest of the population. To guide practical discussions, its truth value could remain largely independent of discrimination and be a constant when encountering data samples from the group. Decisions like this should be made by stakeholders. An example process that gathers stakeholder beliefs across synthetic scenarios and learns a measure of discrimination is presented in Appendix~\ref{synthetic}. 

Aside from truth values, the type of logic (the t-norm and its corresponding evaluation mechanism) also differs across contexts. Choosing a logic can be guided by an intuitive explanation of the nilpotence-based logic characterization, that is, how truth values degrade when multiple uncertain statements hold simultaneously. We recognize the following non-overlapping cases:
\begin{itemize}
    \item[a)] If injecting more uncertain hypotheses would eventually lead to fully erroneous statements (with zero truth values), defer to Lukasiewicz logic. 
    \item[b)] If logical statements are no weaker than their weakest argument, defer to G{\"o}del logic.
    \item[c)] If an original kernel of truth always remains through additional uncertain statements, defer to Product logic.
\end{itemize}
Other criteria or theoretical properties intrinsic to certain logics may also be transcribed to layperson terms and used to make an informed logic selection in place of nilpotence. For example, the truth value of fairness is binary in G{\"o}del and Product logics, meaning that AI systems can either be or fully fail to be fair; there is no intermediate state. In these cases, the continuous $bias$ can be minimized in place of imposing fairness, and for non-zero biases fairness will remain uncertain with truth value less than $1$, and often $0$. Properties like this can also be thought of as restrictions derived from initial choices, but take care to not over-epistemize the logic selection strategy.

\section{Case Studies with Practical Results}\label{demonstration}
We now conduct a series of case studies where our fuzzy logic perspective is not used to create new BL definitions of group bias and fairness, but to critically examine several types of fairness evaluation that are already in use. For the examined evaluation mechanisms, we seek an intuitive understanding of mathematical quantities by matching them to underlying BL expressions that can generate them when enriched with context-specific information. We also introduce variations that apply the same underlying definitions in different contexts. 

We explore four topics: a) the validity of discrimination thresholds in Subsection~\ref{replicating}, b) multidimensional fairness and bias in Subsection~\ref{multi}, c) the ABROCA measure and a variation under Product logic in Subsection~\ref{abroca}, d) the Hooker-Williams criterion for individual fairness and a variation under Lukasiewicz logic in Subsection~\ref{hooker}. 
Our framework and standardization are not limited to these topics, which are chosen to showcase different usages, and may create an intuitive understanding for existing and new fairness evaluation.

\subsection{Replicating base measures of discrimination}\label{replicating}
When discrimination follows Proposition~\ref{discrimination} instead of being directly extrapolated from stakeholder beliefs, it compares base system properties $M(S)$ with truth value $m=\prob(M(S))$ for group membership $S$ against $m'=\prob(M(S'))$ for group membership $S'$. Stakeholders can be engaged in identifying which groups to compare, which system properties matter, or in learning to construct properties from lower-level measurements of reality through. Here we attempt to justify different measures of discrimination that compare AI characteristics between groups within the three base BL subclasses, and create an intuition on thresholding practices, which are widespread but rarely justified.

Table~\ref{tab:discriminations} provides closed-form expressions for the truth value of discrimination (given that it adheres to the above-mentioned proposition) across the three base BL subclasses. The necessary computations are presented in Appendix~\ref{extract formulas}, and future work can define new measures based on the t-norms of other logics. 

To begin with, generic absolute differences $|\Delta m|=|m-m'|$ first introduced in Subsection~
\ref{measures} corresponds to discrimination under Lukasiewicz logic. This logic also justifies the practice of thresholding the differences. For example, to achieve fairness by mitigating worst-case bias it must hold that: 
\begin{align*}
    &0=\prob_L(Bias_{F_{gen}})=e\star_L s=\max\{e+s-1,0\}
    \\&\text{and thus }e\leq 1-s=1-\prob(S)=\prob(\neg S)
\end{align*}
As a result, thresholds can be interpreted as the truth value of \textit{not} encountering group members. This intuition can be communicated to stakeholders to inform the selection of thresholds as a truth value (what ``encountering'' means will differ between contexts). Conversely, if a fairness-aware system admits discrimination $e$, it is fair up to the truth value $1-e$ of not encountering the protected group. 

On the other hand, ratio comparisons fall under Product logic, where bias thresholds can only be artificially integrated through some implication in the predicate $F(S,E)$. For instance, differential fairness already recognizes that additional expressions imposing more conservative definitions of bias are needed \cite{foulds2020intersectional}, whereas counterfactual fairness can satisfy probabilistic formulations up to a small numerical tolerance. One choice that ignores Product logic discrimination less than $\prob(Tol)$ is $$F(S,E)\triangleq\neg \overset{wk}\neg(Tol\to E)$$ with truth value $\prob_\Pi(F(S,E))=\{1\text{ if }e>\prob(Tol),0\text{ otherwise}\}$. This requires a proposition ``$Tol$ implies discrimination'' that has non-zero truth value everywhere for predicate $Tol$. For example, if numerical tolerance of a computed measure of discrimination is the truth value of numerical imprecision, accepting that the statement ``numerical imprecision implies discrimination'' always has some truth would lead to accepting measure values less than the tolerance as fair. Stakeholders may think of cases where measures of discrimination have truth values less than their numerical tolerance, therefore occasionally setting zero truth value for the statement; this concern would invalidate the thresholding strategy.

Finally, setting discrimination as the weak negation of the worst property value is a novel definition that arises from G{\"o}del logic. For example, one minus the worst accuracy between males and females is a measure of discrimination. Under the next section's minimum reduction, generalizing this property to more than two groups means that every group's complement within the population should also be considered as a group. Under G{\"o}del logic, worst case fairness can be achieved for $s>0$ only for perfect values $m=m'=1$.

\begin{table}[htpb]
\small
    \centering
    \begin{tabular}{l c c}
         \textbf{Logic} & $\prob(E)$ & \textbf{Generation space}\\
         \hline
         G{\"o}del & $1-\min\{m,m'\}$ & $e\in[0,1]$\\
         Product$^*$ & $1-\tfrac{\min\{m,m'\}}{\max\{m,m'\}}$ & $e\in[0,1]$\\
         Lukasiewicz & $|m-m'|$ & $e\in[1-s,1]\cup\{0\}$\\
    \end{tabular}
    \caption{Truth values of discrimination $E$ and the generation space under BL subclasses.\\$^*${\scriptsize For $m=m'=0$ Product logic discrimination evaluates to $0$.} }
    \label{tab:discriminations}
\vspace{-1cm}
\end{table}

\subsection{Aggregating multidimensional discrimination}\label{multi}
An emerging consideration for group fairness is that more than one groups could be protected. Group intersections may require further protection due to cumulative effects of discrimination \cite{foulds2020intersectional}. - fairness concerns like these follow a scheme of quantifying discrimination between group pairs or between each group and the total population. They then employ aggregation mechanisms that reduce results to one assessment of bias or fairness \cite{roy2023multi}. In multi-attribute multi-value settings, each value of each attribute becomes a separate group. Multiple fairness constraints are supported by computing multiple measures of discrimination on each group. We aim to write aggregation mechanisms for multidimensional fairness within BL to maintain end-to-end logical consistency.

Rawl's difference principle \cite{ashrafian2023engineering} stipulates that some utility function should be optimized for the worst-of individual. This is a type of individual fairness and can be justified under BL by modeling each individual as a member of a different group, and imposing weak conjunction $\wedge$ across groups. In particular, consider individual biases $Bias_i$ with truth values $bias_i=\prob(Bias_i)$ for individual persons $i=1,\dots,n$ under BL. We define the corresponding fairness term:
\begin{equation}Rawl\triangleq\bigwedge_i\overset{wk}\neg Bias_i\quad\text{for which}\quad\prob(Rawl)=\min_i (1-bias_i)\end{equation}
Weak conjunction is applied incrementally to the biases, and weak negation makes $\overset{wk}\neg Bias_i$ correspond to some idea of utility. Existing interpretations of Rawl's difference principle consider biases to be equal to measures of discrimination, but depending on the logic we can also use other mechanisms.

Now let $i=1,\dots,n$ be groups, subgroups, or intersections thereof \cite{roy2023multi} (these could be overlapping), each stapled with its own fairness concern leading to respective bias and fairness predicates $Bias_i$. We set strong fairness $Fair$ as the conjunction of all fairness definitions, and lack of bias $Unbias$ as the conjunction of all weak negations of bias:
\begin{equation}Fair\triangleq\mathlarger{\mathlarger\&}_i \neg Bias_i,\quad Unbias\triangleq\mathlarger{\mathlarger\&}_i\overset{wk}\neg Bias_i\end{equation}
There is a clear implication from fairness to lack of bias, and to the difference principle applied to groups, namely $Fair\to Unbias\to Rawl$. This usage of the difference principle is atypical but may still form a valid definition in some settings. The three definitions generate reduction mechanisms of varying strength. For example, it holds that $\prob_G(Rawl)=\prob_G(Unbias)=\min_i(1-bias_i)$ in G{\"o}del logic, and $\prob_\Pi(Unbias)=\mathlarger{\mathlarger\Pi}_i(1-bias_i)$ in Product logic. In these two logics, fairness obtains binary truth values, and moving from high but not complete lack of bias to $\prob(Fair)=1$ requires either thresholding via the $F(S,E)$ predicate or exact satisfaction of fairness criteria by imposing them as hard constraints.

\subsection{ABROCA in Product logic}\label{abroca}
Absolute Between-ROC
Area (ABROCA) \cite{gardner2019evaluating} is a measure of fairness across many decision thresholds of binary classifiers. Instead of directly comparing the Area Under Curve (AUC) of such classifiers, this measure of discrimination computes the area between the curves:
\begin{equation}ABROCA\triangleq\int_{0}^1 |tpr(S)-tpr(S')|\,\text{d}fpr\end{equation}
where quantities $tpr(S)$ represent the true positive rate of groups $S$ for the corresponding $fpr$ values. Since true and false positive rates admit probabilistic interpretations, we create an adaptation of this measure under Product logic.

We use the modeling of Yurchenko \cite{yurchenko2021foundations}, which lets Product logic coincide with probability theory and lets us measure the AUC of a group $S$ as the expected truth value $AUC=\mathbb{E}[tpr(S)]=\int_0^1 tpr(S) \,\text{d} fpr$. We now argue that two curves $tpr(S)$ and $tpr(S')$ can be compared in terms of Table~\ref{tab:discriminations}'s Product discrimination instead of via an ad hoc absolute difference. Based on this principle create a measure, called Relative Between-ROC Area (RBROCA), of the expected truth value of discrimination $e$ given measure values $m=tpr(S)$ and $m'=tpr(S')$ between the curves across $fpr$ values. Switching to product logic discrimination from Table~\ref{tab:discriminations} yields:
\begin{equation}RBROCA\triangleq\mathbb{E}[e]=\int_{0}^1 \bigg(1-\frac{\min\{tpr(S),tpr(S')\}}{\max\{tpr(S),tpr(S')\}}\bigg)\,\text{d}fpr\end{equation}

\begin{figure}[htpb]
\centering
\includegraphics[width=0.8\textwidth]{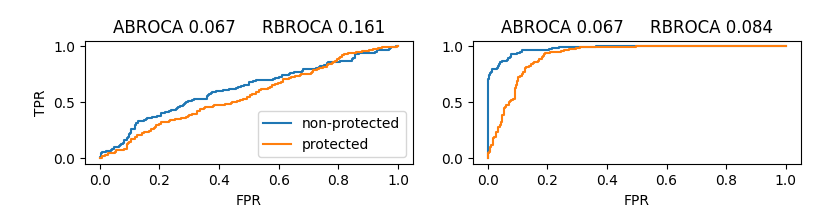}

\caption{Two pairs of different sensitive and non-sensitive attribute ROCs with similar ABROCA but different RBROCA.}\label{avsrbroca}
\end{figure}

Figure~\ref{avsrbroca}  demonstrates this new measure's usefulness in certain cases by presenting two synthetically generated pairs of curves with similar ABROCA. These obtain significantly different RBROCA, as in the second case the curves exhibit a similar shape, whereas in the first case the curves exhibit different slopes. A full series of experiments that qualitative investigate differences between the two measures can be found in Appendix~\ref{abroca comparisons}. We stress that, to our knowledge, no modeling of probability theory can be achieved under Lukasiewicz that would be needed to accomodate the differences between the curves. Thus, RBROCA brings in logical consistency that ABROCA lacks, for example to be conjuncted with other measures when there are multiple fairness criteria.

\subsection{On the Hooker-Williams fairness criterion}\label{hooker}
\citeA{hooker2012combining} introduce a fairness criterion that combines the concepts of equity and utilitarianism. Of these, the former captures notions of individual fairness and the later induces an aggregated objective for the collective instead of individuals. In particular, Rawl's difference principle is traded-off with maximal utility across individuals per:
\begin{equation}hw=(n-1)\delta+n\cdot u_{\min}+\sum_{i}\max\{0,u_i-u_{\min}-\delta\}=-\delta+\sum_{i}\max\{\delta+u_{\min},u_i\}\end{equation}
where $u_{\min}=\min_i \{u_i\}$, $u_i\geq 0$ are utility functions for each individual $i=1,\dots,n$, and the parameter $\delta$ \textit{``is clearly judgemental and likely to be a point of disagreement among the parties concerned''} \cite{hooker2012combining}. This parameter intentionally has the same unit of measurement as the utilities but is still not expressed in intuitive terms. Here we aim to establish it (or a scaled version) as the truth value of some predicate. We also generate a stronger version of the criterion that leverages the fairness reduction of Subsection~\ref{multi} to not presume fairness when there is discrimination against any individual.

Assume that utilities $u_i$ and the value of $\delta$ reside $u_i,\delta+u_{\min},hw\in[0,1]$. For example, they could be scaled to that interval. We can establish $hw$ as a the truth value of an expression that includes fairness definitions. Working under Lukasiewicz logic to let fairness obtain continuous truth values, consider each individual as a group unto themselves. We already pointed out that this last assumption lets group fairness concepts generalize to individual fairness. Then, the equivalent fuzzy logic fairness definition
\begin{equation}HW\triangleq \neg\Delta\,\&\,\neg\mathlarger{\mathlarger\&}_i BiasHW_i\quad \text{where}\quad BiasHW_i\triangleq(\neg\Delta\,\&\, \neg U_{\min})\wedge \neg U_i\end{equation}
has truth value $\prob_L(HW)=hw$ under the selected logic for some predicate $\Delta$ with truth value $\prob_L(\Delta)=\delta$. We consider $BiasHW_i$ as bias definitions because they capture an unfavorable outcome for each individual and can be rewritten in the standard form of Theorem~\ref{Fgroup}:
\begin{equation}\label{biashw}BiasHW_i=E_i\,\&\,S_i\,\&\,F(S_i,E_i)\quad \text{where}\quad E_i=\neg \Delta\,\&\,\neg U_{\min}\text{, }S_i=\neg U_{i}\text{, }F(S,E)=E\to S\end{equation} 

This is a standard form of bias only if Equation~\ref{feasibility} holds, which Appendix~\ref{standardhw} shows to occur only for utilities satisfying $u_{\min}\geq \tfrac{1-\delta}{2}$. Different decompositions, such as ones that that conjunct $F_{gen}$ to $F$, may create different requirements for satisfying the same equation while creating interpretations for $\delta$ other than the following. The above-described decomposition sets $S_i$ as group memberships, which in our setting corresponds to encountering the respective individuals $i$. Individuals with high utility have lesser truth value of being encountered during fairness evaluation, due to utilitarianism disfavoring actions against them. On the other hand, the discrimination predicate $E_i$ has equal truth value among individuals under the assumed equity. Given that $\prob_L(E_i)=1-\delta-u_{\min}$, we get $\delta=1-\prob(E_i)-u_{\min}=\prob(\neg E_i\,\&\,\neg u_{\min})$, that is, the threshold $\delta$ is the truth value of conjuncting non-discrimination against the individual and non-minimum utility. In other words, the threshold holds the \textit{truth value of non-equity for each individual}.

We finally create a variation of HW that adopts the principles of Subsection~\ref{multi}. In Lukasiewicz logic, HW captures the property that non-equity is not implied from any individual being unbiased $HW=\neg(\neg\mathlarger{\mathlarger\&}_i BiasHW_i\to \Delta)$. This is trivial to meet when some individual biases obtain small truth values (e.g., due to small disparity between the greatest and smallest individual utility), therefore creating a near-zero $\prob_L\big(\mathlarger{\mathlarger\&}_i BiasHW_i\big)$ and ultimately saturating $\prob_L(HW)$ to one. Furthermore, small truth values of utility does not satisfy the minimum constraint need for the above-described interpretation of $\delta$; a different interpretation is needed, which can be found Appendix~\ref{standardhw}. We address these shortcoming by considering high (e.g., $\geq 0.5$) utility values for individuals, and stating that lack of bias across all of them does not lead to non-equity. We thus create an expression that we call the Fair Hooker-Williams (FairHW) criterion:
\begin{equation}
    FairHW=\neg \big(\mathlarger{\mathlarger\&}_i \neg BiasHW_i\to \Delta\big)
\end{equation}

\begin{figure}[htpb]
\centering
\includegraphics[width=0.6\textwidth,trim=0px 20px 0 10px,clip]{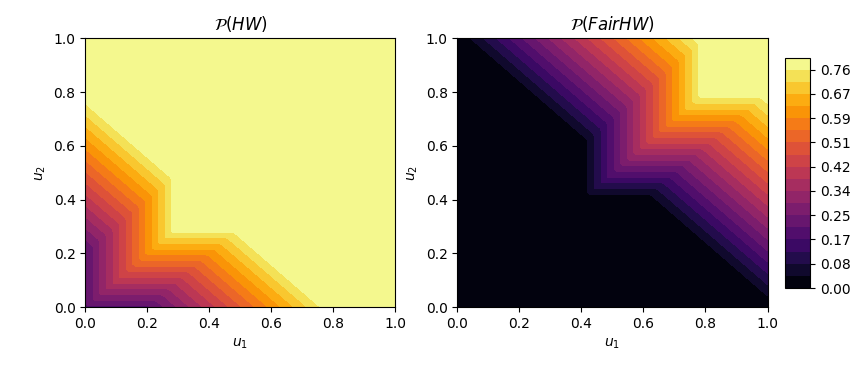}
\caption{Contour of the truth value of HW (left) and FairHW (right), across the utility values of two individuals (axes), and for $\delta=0.2$.}\label{chw}
\end{figure}

In Figure~\ref{chw} we vary the utility of two individuals for a fixed choice $\delta=0.2$. FairHW can capture differences between high utilities, and saturates to $0$ for low ones. HW exhibits the opposite behavior. The original evaluation $hw$ does not saturate anywhere, but is not subject to any logic, and for it the predicate $\delta$ is not interpretable under our modeling.

\section{Discussion}\label{challenges}
The fuzzy logic framework of Figure~\ref{overview} aims to create socially responsible AI \cite{cheng2021socially} by enabling participatory design and democratic deliberation when stakeholder opinions are gathered in each context \cite{weinberg2022rethinking}. This requires an interdisciplinarity collaboration that involves experts from several domains.

First, fuzzy \textit{logic practitioners} or other researchers with the required expertise should accurately express the context-independent part of fairness definitions in BL. These expressions can capture common assumptions, reflect the intent of policy makers, or adhere to regulation. To facilitate direct adoption of our framework by AI system creators in simple scenarios, our analysis in Subsections~\ref{propositions} and~\ref{standardize} removes the need for this transcription in common settings by standardizing the end-result of logical analysis. In this case, AI creators can determine (or set up machine learning systems to determine) how the truth value of the predicate $F(S,E)$ should be computed based on domain expertise, and only the step of encoding stakeholder beliefs remains.

Second, \textit{social scientists} should gather stakeholder feedback. Among stakeholders, policy makers may also be the ones contriving BL definitions. For example, they may be legal experts operationalizing regulation, though familiarization with BL can be arduous without prior experience. In the simplest scenario, policymakers would work in tandem with logic practitioners, similarly to the collaborative processes with which causal models are built from domain expertise. Gathering stakeholder feedback needed by our framework also presents challenges other than gathering and parsing potentially conflicting viewpoints. For instance, there could be discussions on how predicates being quantified should be interpreted, as well as a negotiation of which criteria they would be quantified under. In this regard, quantitative feedback may be harder to gather than qualitative discussions, and the burden of a conversion may fall on social scientists gathering it, with accompanying subjectivity concerns. Questionnaires or other means of extracting beliefs may also need interdisciplinary design, for example to account for edge cases both algorithmically and socially. 

Third, \textit{computer scientists} (e.g., data analysts, machine learning engineers) are needed to implement the extracted definitions of fairness or bias. These are primarily the AI creators that should drive the adoption of this work in practice.

Fostering a collaboration between the above people presents its own challenges, ranging from organizational complexity to reconciling different viewpoints (e.g., sociological vs technical). The collaboration is at worst as difficult as that of other interdisciplinary fairness approaches, compared to which ours creates reusable BL definitions that reduce future efforts and, importantly, lets us understand and adjust algorithmic practices. 

\begin{figure}[htpb]
\centering
\includegraphics[width=0.65\textwidth]{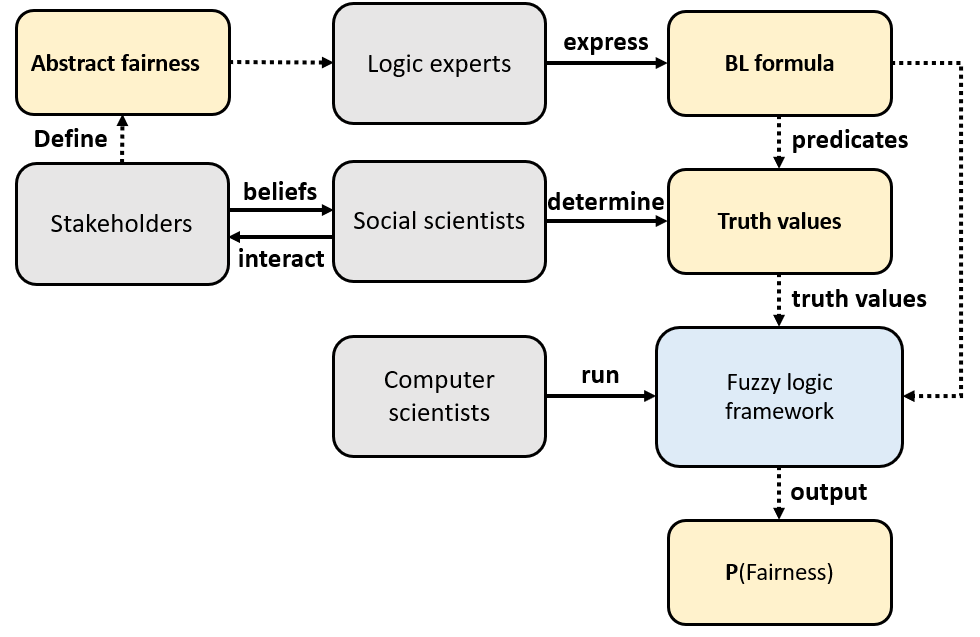}
\caption{Interdisciplinary collaboration to apply our framework in practice.}\label{roles}
\end{figure}

One prospective scheme that reduces the interactions between disciplines is presented in Figure~\ref{roles}. In this, logic experts are tasked with creating BL formulas, social scientists with determining truth values, and computer scientists with running the fuzzy logic framework. Computer scientists and logic experts could be the same people. Both logic experts and social scientists gather the context-independent and context-specific opinions of stakeholders, respectively encapsulated in abstract fairness definitions and in truth values. Identifying the minimum number of data points (e.g., the number of questionnaire respondents) given the complexity of learning models warrants further research, as it can help budget the generation of context-dependent variations of previously stated BL definitions.

Throughout this work, we provide information that could be useful to all the roles, from theoretical details to transcription of several decisions to layperson terms. However, we focus foremost on theoretical facets to establish our framework on a solid basis. Further refinement may be needed before practical adoption. For example, other propositions may be explored, our methodology may be enriched with detailed guides, and additional technical research can simplify or improve parts of fairness definition and evaluation pipelines. Large language models can also help transcribe natural language stipulations in BL. A final challenge is that more complex predicates may be needed to overcome expressive limitations of fuzzy logic, such as to prevent implication between independent predicates. However, this additional groundwork only arises when generating BL expressions and not during context-specific adaptation.

\section{Conclusions and Future Work}\label{conclusions}
In this work we introduced the concept of stating predicate-based definitions of group fairness within BL and creating numerical evaluation of bias and fairness by accounting for context-specific stakeholder input. To support practical adoption of this framework, we theoretically explored common propositions of what constitutes group fairness, and provided standardized forms of predicate truth values. We finally showcased results on four fairness topics, where we were able to provide intuitive predicate-based interpretation of existing practices and proposed adaptations in new contexts that demonstrably maintain logical consistency.

Our framework is not limited to this work's demonstrations and could grant intuitive understanding across a wide range of fairness definitions during during practical adoption to engage stakeholders. Further research is needed on how belief systems can be extracted from real persons through social processes or causal modeling. One interesting interdiscriplinary direction would be to transcribe emerging fairness laws in BL and let their interpretation be guided by context-specific beliefs. We are also interested in exploring counterfactual fairness with world models evaluated in subclasses of BL.

\section*{Acknowledgements}
This work was funded by the European Union under the Horizon Europe projects ELIAS, Grant Agreement ID: 101120237, and MAMMOth, Grant Agreement ID: 101070285. UK participant in the Horizon Europe Project MAMMOth is supported by UKRI grant number 10041914 (Trilateral Research LTD).

\appendix

\section{Fuzzy Logic Theory}

\subsection{T-norms and residuum}\label{tnorms}
MTL logics $\mathcal{L}$ enlist modus ponens as the only deduction rule, i.e., from $X$ and $X\to Y$ infer $Y$ and define evaluation mechanisms $\mathcal{P}_{\mathcal{L}}:\mathcal{L}\to [0,1]$ that produce truth values $x=\mathcal{P}_{\mathcal{L}}(X)$ for statements $X\in\mathcal{L}$. These evaluation mechanisms are based on t-norms, which are operations $\star_\mathcal{L}:[0,1]^2\to[0,1]$ that quantify the truth value of logical conjunction. That is: $$\mathcal{P}_\mathcal{L}(X\& Y)=\mathcal{P}_\mathcal{L}(X)\star_\mathcal{L} \mathcal{P}_\mathcal{L}(X)$$ We reiterate that $\star$ is used when stating properties for any logic. T-norms are left-continuous and satisfy the following properties for truth values $x,y,z\in [0,1]$ of some predicates:
\begin{center}
$x\star 0=0$, $x\star 1=1$, $x\star y=y\star x$, $(x\star y)\star z=x\star(y\star z)$, if $x\leq y$ then $(x*z)\leq (y*z)$
\end{center}
In MTLs, conjunction is not always idempotent in that it may hold that $\mathcal{P}(X)<\mathcal{P}(X\& X)$ for some $X$. Implication is evaluated via the t-norm's residuum operator $\Rightarrow$, that is:
$$\mathcal{P}(X\to Y)=\big(\mathcal{P}(X)\Rightarrow\mathcal{P}(Y)\big)$$ for statements $X,Y$, where the residuum is in turn defined as:\
$$(x\star y)\leq z\triangleq  x\leq (y\Rightarrow z)$$
Substituting $x=1$ in the above formula, we can see that $y\Rightarrow z$ holds strictly (with truth value $1$) only if $y\leq z$, which is a property we use extensively in this work.

\subsection{Fuzzy Logic Axiomatizations}\label{axiomatization}
For the sake of completeness, we now present the Hilbert-like proof system for various MTLs. First, they all admit the following axioms:
\begin{itemize}
    \item[] MTL1: $(X\to Y)\to ((X\to Z)\to(Y\to Z))$
    \item[] MTL2: $X\,\&\,Y\to X$
    \item[] MTL3: $X\,\&\,Y\to Y\,\&\,X$
    \item[] MTL4a: $X\wedge Y\to X$
    \item[] MTL4b: $X\wedge Y\to Y\wedge X$
    \item[] MTL4c: $X\,\&\,(X\to Y)\to X\wedge Y$
    \item[] MTL5a: $X\to(Y\to Z)\to ((X\,\&\, Y)\to Z)$
    \item[] MTL5b: $((X\,\&\, Y)\to Z)\to X\to(Y\to Z)$
    \item[] MTL6: $((X\to Y)\to Z)\to ((Y\to X)\to Z)\to Z)$
    \item[] MTL7: $False\to X$
\end{itemize}
This system is extended to BL by adding the following axiom:
\begin{itemize}
    \item[] BL: $X\,\&\,(X\to Y)\to Y\,\&\,(Y\to X)$
\end{itemize}

In table~\ref{tab:axioms} we now show how the base MTL logic can be extended with additional axioms to model the three base BL subclasses. That is, reasoning under each logic's axioms yields the same outcome as if computing the truth values of predicates via corresponding t-norms and residuums. Throughout this work, we defer to the later practice, but analytical computations are theoretically corroborated by axiomatization and the logical rule of modus ponens. In general, proofs in fuzzy logic may be more complex or counter-intuitive than in propositional logic, but the outcome of this work only uses the axiomatization to \textit{state} expressions of fairness and bias; any inference of what these imply is handled by our analysis.

\begin{table}[htpb]
    \centering
    \begin{tabular}{l l}
         \textbf{Logic} & \textbf{Additional axiom}\\
         Godel & $X\to X\,\&\,X$ \\
         Product & BL and $\neg X\vee((X\to X\,\&\,Y)\to Y)$ \\
         Lukasiewicz & $((X\to Y)\to Y)\to ((Y\to X)\to X)$
    \end{tabular}
    \caption{Axiomatic extension of MTL to base BL subclasses.}
    \label{tab:axioms}
\end{table}

Each t-norm admits two types of negation: a) the older fuzzy set negation predicate that is added on top of MTLs, which we call weak negation, and b) the more recently identified axiomatic negation statement of the logic, which we call strong negation. The truth value of these negations are respectively defined per:
$$\overset{wk}\sim x\triangleq  1-x,\quad \overset{str}\sim x\triangleq  (x\Rightarrow 0)$$
Weak negation always satisfies that double negative yields the original truth value $\overset{wk}\sim\overset{wk}\sim x=x$, and therefore can interpret fuzzy set theory. We annotate it as:$$\prob(\overset{wk}\neg X)=\overset{wk}{\sim}\prob(X)$$ On the other hand, strong negation is an in-built construct of MTLs and satisfies that conjunction with a statement's negative implies a fully erroneous statement:$$\big(x\star(\overset{str}\sim x)\big)=\big(x\star(x\Rightarrow 0)\big)\Rightarrow \big(0\star(0\Rightarrow x)\big)=0$$
We annotate logical statements using strong negation with the vanilla negation symbol:$$\prob(\neg X)=\prob(X\to False)=\big(\prob(X)\Rightarrow 0\big)=\overset{str}{\sim}\prob(X)$$
where $False$ is the logic's fallacy predicate with $\prob(False)=0$. All MTLs can model fuzzy set logic (the logic with weak conjunction and negation) and are essentially extended with additional axioms corresponding to the choice of t-norms.

In terms of creating more complex definitions of fairness in BL, we point that first-order logic quantifiers ($\forall$, $\exists$) can also be modelled within any MTL $\mathcal{L}$, such as in $\mathcal{L}=BL$. In this work we consider this extension (it also admits the corresponding axiomatic modeling) by stipulating the truth value of the generalization quantifier using weak conjunction:
\begin{equation}
    \forall X\,Y\triangleq\bigwedge_{X\in\mathcal{L}} Y\quad\text{with truth value}\quad \prob(\forall X\,Y)=\inf_{X\in \mathcal{L}}\prob(Y)
\end{equation}
and defining the existence quantifier per $\exists X\,Y\triangleq \neg(\forall X\,\neg Y)$. 


\section{Proofs}\label{proofs}

\noindent\textsc{Theorem~\ref{exists imbalance}}. \textit{For every imbalance definition $Imb$ there exists expression $F(S,E)$ such that:
\begin{equation*}
    Imb_{F}\triangleq S\to (S\,\&\,E\,\&\,F(S,E))
\end{equation*}
satisfies $s\star\prob(Imb_{F})=s\star imb$ as long as: 
\begin{equation*}
    \prob(F(S,E))\leq \prob(F_{gen}(S,E))\text{ where }\prob(F_{gen}(S,E))\triangleq\Big(e\Rightarrow \inf_{s\star d\geq s\star e} d\Big)
\end{equation*}}
\begin{proof} 
For $s=0$ we obtain $\prob(Imb_F)=(0\Rightarrow (0\star e\star \prob(F(S,E))))=(0\Rightarrow 0)=1$.\\For $s>0$ let us define some predicate:
\begin{equation*}
    F(S,E)= F_{gen}(S,E)\,\&\,F_{repl}(S,E)\,\text{where}\,\prob(F_{gen}(S,E))=(e\Rightarrow e_{\inf})\,\text{and}\,e_{\inf}=\inf_{s\star d\geq s\star e}d
\end{equation*}
This satisfies $\prob(F(S,E))\leq \prob(F_{gen}(S,E))$ and we will now show that that its truth value exists by showing that there exists truth value for the helper predicate $f_{repl}=\prob(F_{repl}(S,E))$ for which it holds that $s\star\prob(Imb_{F})=s\star imb$. To see this, start from the simplification:
\begin{align*}
    s\star\prob(Imb_F)&=s\star(s\Rightarrow (s\star e\star(e\Rightarrow e_{\inf})\star f_{repl}))
    \\&=\min\{s,s\star e\star(e\Rightarrow e_{\inf})\star f_{repl}\}
    \\&=s\star e\star(e\Rightarrow e_{\inf}) \star f_{repl}
    \\&=s\star \min\{e,e_{\inf}\}\star f_{repl}
    \\&=s\star e_{\inf}\star f_{repl}
\end{align*}
Now let us consider the two edge cases for the helper predicate:\\
\noindent- For $F_{repl}(S,E)=True$ it holds $s\star\prob(Imb_F)=s\star e_{\inf}$.
\\- For $F_{repl}(S,E)=False$ it holds $s\star\prob(Imb_F)=0$. 
\\Write $s\star\prob(Imb_F)$ as a function of the helper predicate's truth value $f_{repl}$. This function is continuous for any continuous t-norm and thus admits some $f_{repl}\in[0,1]$ to yield any value $s\star \prob(Imb_F)\in[0,s\star e_{\inf}]$. Since it also holds $s\star imb\leq s\star e=s\star e_{\inf}$, we thus obtain that some $f_{repl}$ can yield $s\star \prob(Imb_F)=s\star imb$.
\end{proof}

\noindent\textsc{Theorem~\ref{Fgroup}}. \textit{For a standardized form of imbalance $Imb_F$, the corresponding group bias definition has truth value:
$$bias_F=s\star e\star\prob(F(S,E))$$}
\begin{proof}
    \begin{align*}
        bias_F&=\prob(S\,\&\,Imb_F)
        \\&=\prob\big(S\wedge (S\,\&\,E\,\&\,F(S,E))\big)
        \\&=\min\{s,s\star e\star \prob(F(S,E))\}
        \\&=s\star e\star \prob(F(S,E))
    \end{align*}
\end{proof}

\noindent\textsc{Theorem~\ref{worst bias}}. \textit{Any bias definition in BL has truth value at most:
\begin{equation*}bias\leq bias_{F_{gen}}=s\star e\end{equation*}}
\begin{proof}
For $s=0$, the equality holds true by the imbalance definition (all biases have truth value $1$ in this case). Otherwise, from the proof of Theorem~\ref{exists imbalance} above, we obtain the formula and truth value of the predicate $F_{gen}(S,E)$, for which $\prob(Imb_{F_{gen}})=s\star e\star \prob(F_{gen}(S,E))=s\star e_{\inf}$. However, during the same proof we showed that $s\star e=s\star e_{\inf}$, which means that $\prob(Bias_{F_{gen}})=s\star e$. This holds for any pair of values $e,s$. Since $imb\leq e$ and t-norms are non-decreasing, we also obtain that $bias\leq s\star e$.
\end{proof}

\section{Example Questionnaire}\label{synthetic}
Here, we outline a simple methodology for extracting context-specific stakeholder beliefs that let us evaluate an example definition of fairness under the framework of Figure~\ref{overview}. The methodology presents intuitive explanations to stakeholders and thus supports more a pluralistic input than enlisting domain expert opinions only in defining fairness. We show numeric results using a synthesized collection of example stakeholder answers and negotiation outcomes.

~\\{\textsc{Preliminary: Fairness Definition in BL}}\\
In general, definitions of fairness will not be too complicated, as we want them to be intuitive and highly interpretable. If there are no other propositions to be modelled, worst-case fairness should suffice. We expect more complex forms to arise when there are several definitions to be simultaneously assessed, or when there are complex propositions to conjunct. For the sake of demonstration, let us consider a standard form definition:
$$Fair=(S\star E\star F(S,E)\to False)$$
with a twist of introducing a ``business necessity'' predicate $B$ that, when it occurs for group members, should never imply discrimination, i.e., $((B\star S)\to E)\to False$. This could be part of affirmative action that hires group members at a higher position, though keep in mind that introducing too complicated clauses may have unforeseen shortcomings, as happens in this case below. We add this property (not the predicate) in the definition by conjuncting it to $F_{gen}(S,E)$ to construct $F(S,E)$. That is, we create the following toy expression:
$$F(S,E)=F_{gen}(S,E)\,\&\,(((B\star S)\to E)\to False)$$
We now show what stakeholder feedback would look like, and how it would let us evaluate this definition based on context-specific beliefs.\\

\begin{figure}[htpb]
\centering
\includegraphics[width=0.7\textwidth]{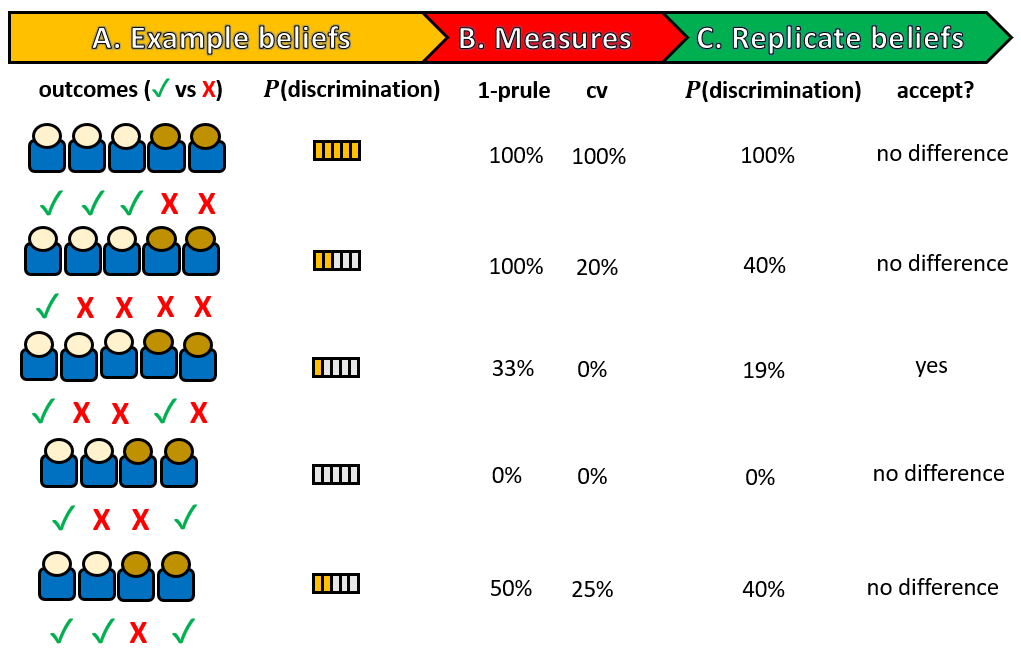}
\caption{Synethetic example of fitting survey-derived discrimination.}\label{belief}
\end{figure}

\noindent{\textsc{A. Example beliefs}}\\
As a first step, we provide example AI system outcomes to stakeholders so that they encounter a wide breadth of situations across the real-world context in which systems are expected to operate. For each set of outcomes, we ask them to evaluate discrimination in the range $[0,1]$. This corresponds to obtaining the \textit{truth value} of discrimination in reference cases. Recall that these are not the truth values of bias, and therefore stakeholders need not worry about reasoning on whether their observed discrimination is acceptable. A running example on measuring race discrimination for a hiring process is presented in Figure~\ref{belief}; this is artificially constructed for demonstration purposes and does not replicate any real-world data or system.\\

\noindent{\textsc{B. Measures}}\\
We assume that an axiomatic definition of discrimination is not possible in this setting, for example because stakeholders cannot agree on which predictive properties should be compared between groups. We instead determine which quantitative data properties loosely replicate similar principles as the ones that influence beliefs, for example through workshops involving domain bias experts and stakeholders. 
In the running example, discrimination is considered to pertain to some (yet partially unknown) combination of disproportional and unequal representation in hiring results. These correspond to the $1-prule$ and $cv$ measures, though others could be used too.\\

\noindent{\textsc{C. Replicate beliefs}}\\
We now create a mechanism that replicates the truth values of discrimination. In the simplest case, this could be an existing measure found the literature, or a machine learning model trained on the gathered truth values of discrimination. However, more complicated mechanisms that combine the quantities pinpointed in the previous step could be created. In the running example, the measures are combined through a two-layer perceptron neural network with $64$ hidden layers, and trained with gradient descent with $10^{-3}$ learning rate on the five examples until numerical tolerance $10^{-6}$. There will likely be small deviations from the reference truth values of discrimination on validation data (for simplicity, in the running example training data are also used for validation). It is important to again involve stakeholders and policymakers in determining which deviations are acceptable, as shown in the last column of the running example. We could this or the previous step if some deviations were not acceptable, for example to retrain the two-layer neural network to greater precision, or to rework which quantities it should learn from. The end-result of the above process is a measure of discrimination (or a model approximating such a measure) that mirrors the belief systems of stakeholders. 

\noindent{\textsc{D. Determine remaining choices}}\\
We finally need to decide on what the group membership predicate $S$ entails and its corresponding truth value, as well as in which subclass of BL fairness evaluation will take place. In the specific definition of fairness we tackle in this appendix, we also need to determine the truth value of the business necessity predicate $B$. For the first property, we consider a scenario where interaction with stakeholders led to consider the truth value of encountering one group members as the statistical chance of encountering a person from the group in the real world. This value could be the result of negotiation, but could also be obtained through an additional questionnaire. Let us assume that some hiring policy also set the business necessity predicate to be equal to $10\%$. To determine which fuzzy logics we should work with, refer to nilpotency as a criterion already explained in Subsection~\ref{logictype}; we ask the same questions as that subsection to stakeholders. For example, let us assume that, after further negotiation, preserving some minimum kernel of truth value was most important for them. This means that the fairness definition is evaluated in Godel logic. For this, $\prob_G(F_{gen}(S,E))=1$.

~\\\noindent{\textsc{Numerical example}}\\
We finally present an example evaluation of our toy fairness definition: for a set of predictions with $prule=77\%$, $cv=0\%$, business necessity $30\%$ and which covers $s=5\%$ of the population. For these values, our neural network predicts discrimination $e=19\%$, and therefore
\begin{align*}
    \prob_G(Fair)&=(19\%\star_G 5\%\star_G 1\star_G((5\%\star_G 10\%\Rightarrow 19\%)\Rightarrow 0))\Rightarrow 0
    \\&=(5\%\star_G((5\%\Rightarrow 19\%)\Rightarrow 0))\Rightarrow 0
    \\&=1
\end{align*}
After this first evaluation, we get a sense that the business necessity predicate may be highly contentious; any value $\prob_G(B)>0$ yields $\prob_G(Fair)=1$ for $s>0,e>0$. To avoid similar pitfalls, we stress that properties of fairness definitions written in BL should be critically examined. Furthermore, notice how uninformative working with strong negation may be in Godel or Product logic; although it makes it possible to procure perfectly certain fairness, in practice it does so by discounting a wide range of computational nuance. Thus, it may be preferably to create continuous evaluation by working with bias or its weak negation and only assess fairness at the very end of algorithmic bias mitigation pipelines.

\section{Properties of Base BL Subclasses}\label{extract formulas}
Here, we compute the truth value of discrimination in Equation~\ref{discform}, as well as the generation space sufficient to set $\prob(F(S,E))\leq 1$ in Equation~\ref{feasibility} for the three base BL subclasses, namely Godel, Product, and Lukasiewicz logic. We do so by substituting the t-norm connectives of those logics in the equations, retrieved from Table~\ref{tab:t-norms}. Recall that, in each logic, we annotate $m=\prob(M(S))$, $m'=\prob(M(S'))$. The results of this appendix were previously summarized in Table~\ref{tab:discriminations}.

~\\\noindent{\textsc{Godel discrimination}}
\begin{align*}
    &\prob_G(E)
    =\overset{wk}{\sim}((m\to_G m')\*(m'\to_G m))
    \\&=1-\min\{\{1\text{ if }m\leq m',m'\text{ otherwise}\}, \{1\text{ if }m'\leq m,m\text{ otherwise}\}\}
    \\&=1-\min\{m',m\}
\end{align*}

\noindent{\textsc{Godel generation space}}\\
- When $e<s$ it holds that $\inf_{\min\{s,d\}\geq \min\{s,e\}}d=\inf_{\min\{s,d\}\geq e}d=e$.\\
- When $e\geq s$ it holds that $\inf_{\min\{s,d\}\geq \min\{s,e\}}d=\inf_{\min\{s,d\}\geq s}d=s$.\\
As a result, $1=(e\Rightarrow_G \inf_{s\star_G d\geq s\star_G e}d)$ holds only if $e\geq \inf_{s\star_G d\geq s\star_G e}d=\inf_{\min\{s,d\}\geq \min\{s,e\}}d$
$=\inf_{\min\{s,d\}\geq \min\{s,e\}}d$, which holds for both $e<s$, where $e\geq e$, and for $e\geq s$, where we obtain the requirement $e\geq s$. Therefore, the generation space for Godel logic is $e\in[0,1]$.

~\\\noindent{\textsc{Product discrimination}}
\begin{align*}
    &\prob_\Pi(E)
    =\overset{wk}{\sim}((m\to_\Pi m')\*(m'\to_\Pi m))
    \\&=1-\min\{(1\text{ if }m\leq m',\tfrac{m'}{m}\text{ otherwise}), (1\text{ if }m'\leq m,\tfrac{m}{m'}\text{ otherwise})\}
    \\&=1-\min\{\min\{1, \tfrac{m}{m'}\}, \min\{1, \tfrac{m'}{m}\}\}
    =1-\min\{\tfrac{m}{m'}, \tfrac{m'}{m}\}
    =1-\frac{\min\{m',m\}}{\max\{m',m\}}
\end{align*}

\noindent{\textsc{Product generation space}}\\
$1=(e\Rightarrow_\Pi \inf_{s\star_\Pi d\geq s\star_\Pi e}d)$ holds only if $e\geq \inf_{s\star_\Pi d\geq s\star_\Pi e}d=\inf_{s\cdot d\geq s\cdot e}d=\inf_{d\in[e,1]d}=e$. Therefore, the condition holds true for all $s,e\in[0,1]$.

~\\\noindent\textsc{Lukasiewicz discrimination}
\begin{align*}
    &\prob_L(E)
    =\overset{wk}{\sim}((m\to_L m')\*(m'\to_L m))
    \\&=1-\min\{\min\{1,1-m+m'\}, \min\{1, 1-m'+m\}\}
    \\&=1-\min\{1-m+m', 1-m'+m\}
    =\max\{-m+m',-m'+m\}
    =|m-m'|
\end{align*}

\noindent{\textsc{Lukasiewicz generation space}}\\
- When $e\geq 1-s$ it holds that $\max\{d+e-1,0\}\geq\max\{s+e-1,0\}=s+e-1$ only for $d\geq e$. Thus $\inf_{s\star_L d\geq s\star_L e}d=e$.\\
- When $e< 1-s$ it holds that $\max\{d+e-1,0\}\geq\max\{s+e-1,0\}=0$ for any $d\in[0,1-s]$. Thus $\inf_{s\star_L d\geq s\star_L e}d=0$.\\
As a result, $1=(e\Rightarrow_L \inf_{s\star_L d\geq s\star_L e}d)$ holds only if $e\geq \inf_{s\star_L d\geq s\star_L e}d$, which holds only in the case $e\geq 1-s$ or $e=0$.

\section{Comparison of ROC-based Product Biases}\label{abroca comparisons}
In this appendix we run a series of controlled experiments that let us spot evaluation differences between ABROCA and RBROCA. To this end, we construct two families of synthetic predictions and corresponding predictive scores with controlled AUC. The two families emulate evaluation thresholds for a protected and non-protected population subgroup respectively. Since AUC is largely independent on the fraction of positive labels, we set ground truth labels in each family to be sampled from values $y_1,y_2\in\{0,1\}$ with equal probability. For each ground truth label, we create a corresponding prediction score:
$$\hat{y}_1\in (1-y_1)U_{[0,1]}^c+y_1 U_{[0,1]},\quad \hat{y}_2\in (1-y_2)U_{[0,1]}^c+y_2 (1-U_{[0,1]}^c)$$
where $U_{[0,1]}$ refers to values uniformly sampled from the range $[0,1]$ and the exponent $c\geq 1$ lets us control the expected AUC values of the ROC for each set of predictions. In particular, denoting some AUC scores that would be obtained for each of the curves as $AUC_1$ and $AUC_2$ respectively, these have expected values $\mathbb{E}[AUC_1]=\mathbb{E}[AUC_2]=0.5$ for $c=1$, given that for both curves exhibit an 50\% chance of assigning greater prediction scores for the positive ground truth labels for that parameter value. Furthermore, $\lim_{c\to\infty}\mathbb{E}[AUC_1]=\lim_{c\to\infty}\mathbb{E}[AUC_2]=1$ given that $\lim_{c\to\infty}\tfrac{\hat{y}_1(1-y_1)}{\hat{y}_1y_1}=\lim_{c\to\infty}\tfrac{\hat{y}_2(1-y_2)}{\hat{y}_2y_2}=0$.

In the end, we construct the two families $r_1=\{r_{11},r_{12},r_{13},\dots\}$ and $r_2=\{r_{21},r_{22},r_{23},\dots\}$ of prediction sets $r_{1i},r_{2i}$ by varying $c\in[1,2^5]$, and compute their AUC scores. Each prediction within every $r_{1i}$ or $r_{2i}$ is accompanied with the corresponding ``ground truth'' label $y_1,y_2$ used in its creation. We also construct each pair of curves $(r_{1i},r_{2i})$ using the same seed across different $i$, in order to obtain smoother AUC transitions across different $c$. 

\begin{figure}[htpb]
\centering
\includegraphics[width=0.3\textwidth]{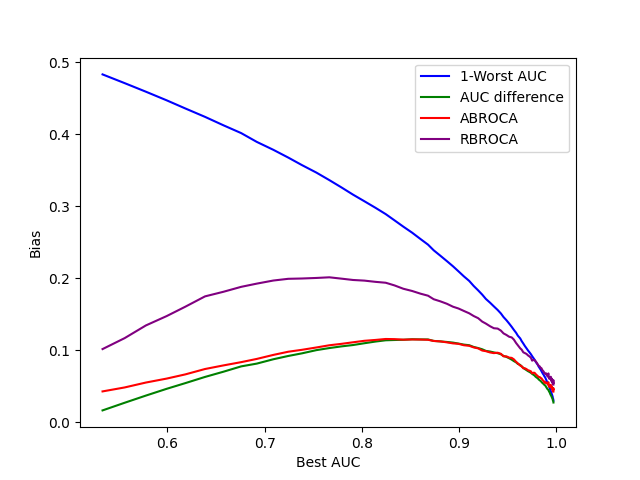}
\includegraphics[width=0.3\textwidth]{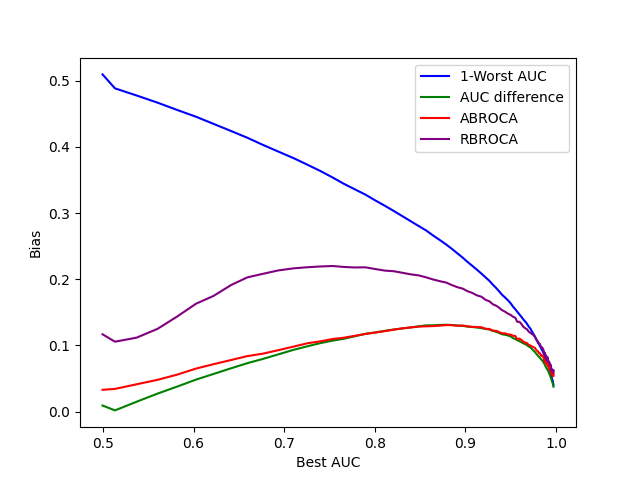}
\includegraphics[width=0.3\textwidth]{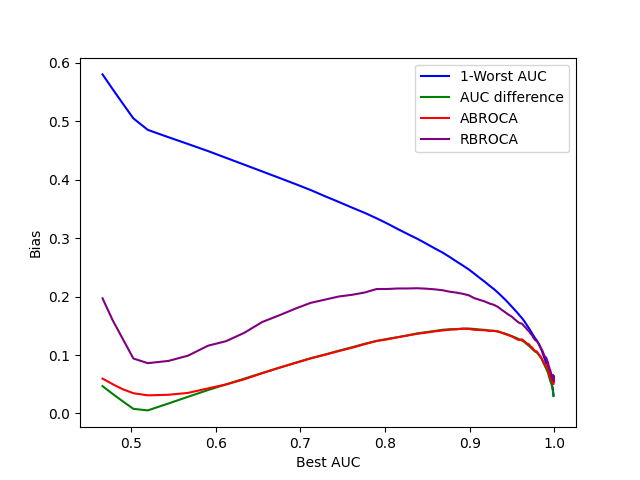}
\caption{Comparison of ABROCA and RBROCA scores over pairs of sensitive and non-sensitive attribute ROCs. Each subfigure repeats the series of experiments with a different randomization seed.}\label{allbroca}
\end{figure}

In Figure~\ref{allbroca} we compare the AUC scores between the two families of curves under different measures of discrimination (vertical axis); the difference of AUCs, ABROCA, RBROCA, and $1-$ the worst AUC for each pair of curves (the last measure would be obtained from Godel logic discrimination for the AUC property). We make these computations across pairs of curves $(r_{1i},r_{2i})$ for the point in the horizontal axis that represents $\min\{AUC(r_{1i}),AUC(r_{2i})\}$. The whole process is repeated for three different randomization seeds.

As a general trend, and for the specific series of experiments we run, ABROCA closely follows the difference of AUCs. This is an expected behavior given that the first family of synthetic predictions converges a little more slowly (as $c$ increases) towards perfect classification for any threshold compared to the second one, and thus its ROCs tend to lie underneath the latter's. However, RBROCA manages to find more substantial differences between the curves, exhibiting inflection points at different values. This corroborates its viability as an alternative measure of discrimination that performs a different type of evaluation.

\section{Standard Form of the Hooker-Williams Bias Term}\label{standardhw}
\noindent\textsc{Standard form requirement}\\
In this appendix we will show that the decomposition presented in Equation~\ref{biashw}, is a standard form of bias by adhering to Equation~\ref{feasibility}. This particular decomposition is used in the text to obtain an intuitive explanation of the parameter $\delta$. For our modeling to be a valid standard form of bias, and denoting for the sake of brevity $e=\prob_L(E_i),s=\prob_L(S),f=\prob_L(F(S,E)),u_{\min}=prob_L(U_{\min}),u=\prob_L(U_i),\delta=\prob_L(\Delta)$ with $u\geq u_{\min}$, we must satisfy:
$$(s\Rightarrow_L e)\leq (e\Rightarrow_L (\inf_{s\star_L d\geq s\star_L e}d))$$
From the generation space analysis, we know that for $e\geq 1-s$ or $e=0$ it always holds $(s\Rightarrow_L e)\leq 1$, and for $0<e<1-s$ we obtain that $(e\Rightarrow_L s)\leq 0$, which holds only for $1-e+s\leq 0$.
When $e=0$ we allow completely certain utility. We now explore which truth values satisfy $e> 1-s$ by rewriting it to
$\max\{1-\delta-u_{\min}, 0\}> 1-(1-u)$, which is equivalent to:
$\delta> 1-u_{\min}-\sup_{u:u<1}u$. However, the equation can only accept utilities whose pairwise sums remain less than $1-\delta$, and thus the proposed decomposition is not a standard form of bias in scenarios where high utilities are involved. We instead explore which truth values satisfy $e\leq 1-s$ by rewriting it to $\max\{1-\delta-u_{\min}, 0\}\leq 1-(1-u)=u$. This can only be satisfied for every individual if: $$u_{\min}\geq \tfrac{1-\delta}{2}$$

\noindent\textsc{Infeasibility of alternative decomposition}\\
A more intuitive interpretation would be to exchange the interepretation of $E_i$ and $S_i$, so as to make the group membership function $S_i$ independent of data samples. This has all the formulaic characteristics of a standard bias form: 
\begin{equation*}BiasHW_i=S_i\,\&\,E_i,\&\,F(S_i,E_i)\quad \text{where}\quad S_i=\neg \Delta\,\&\,\neg U_{\min}\text{, }E_i=\neg U_{i}\text{, }F(S,E)=S\to E\end{equation*}
However, this last decomposition in terms is \textit{not} a bias standard form, due to violating Equation~\ref{feasibility}. In particular, in this case, for $0<e<1-s$ we obtain that $(s\Rightarrow_L e)\leq 0$, which holds only for the contradictory $0\geq 1-s+e>2-2s>0$. As a result, the only viable interpretation would be for $e=0$ and $e\geq 1-s$, which can only be satisfied if $1-u> 1-\max\{1-\delta-u_{\min}, 0\}$, and equivalently: $$\delta\geq 1-u_{\min}-\sup_{u:u<1}u$$
We can therefore see that allowing utilities with value greater than $0.5$ (or, in general, that could sum to a quantity of 1 or greater) prevents the existence of any $\delta$ and therefore also makes it impossible to satisfy Equation~\ref{feasibility}. Intuitively, this means that there exists no underlying imbalance formulation that contracts the declared measure of discrimination.

\vskip 0.2in
\bibliographystyle{theapa}
\bibliography{main}

\end{document}